\documentclass[11pt,a4paper]{article}
\usepackage[hyperref]{acl2017}
\usepackage{times}
\usepackage{latexsym}
\usepackage{graphicx} 
\usepackage{algorithm}
\usepackage{algorithmic}
\usepackage{url}
\usepackage[procnames]{listings}
\usepackage{color}
\usepackage{latexsym}
\usepackage{amsmath}
\usepackage{verbatim}
\usepackage{amssymb}
\usepackage{amsthm}

\usepackage{booktabs} 
\usepackage{multirow}

\aclfinalcopy 
\def\aclpaperid{555} 

\newcommand{\vect}[1]{\mathbf{#1}}
\newcommand{\set}[1]{\mathcal{#1}}
\newcommand{\struct}[1]{\boldsymbol{#1}}
\newcommand{\softmax}{\mathrm{softmax}}

\newcommand{\ignore}[1]{}

\newcommand{\dycomment}[1]{\textcolor{red}{\bf \small [#1 --DY]}}

\newenvironment{itemizesquish}{\begin{list}{\setcounter{enumi}{0}\labelitemi}{\setlength{\itemsep}{-0.25em}\setlength{\labelwidth}{0.5em}\setlength{\leftmargin}{\labelwidth}\addtolength{\leftmargin}{\labelsep}}}{\end{list}}

\title{Program Induction by Rationale Generation:\\
Learning to Solve and Explain Algebraic Word Problems}

\author{
Wang Ling$^{\spadesuit}$ \qquad  Dani Yogatama$^{\spadesuit}$ \qquad Chris Dyer$^{\spadesuit}$ \qquad Phil Blunsom$^{\spadesuit\diamondsuit}$ \\
$\spadesuit$DeepMind$ \qquad \diamondsuit$University of Oxford\\
{\tt \{lingwang,dyogatama,cdyer,pblunsom\}@google.com}
}

\begin{document}

\maketitle

\begin{abstract}
Solving algebraic word problems requires executing a series of arithmetic operations---a program---to obtain a final answer. However, since programs can be arbitrarily complicated, inducing them directly from question-answer pairs is a formidable challenge. To make this task more feasible, we solve these problems by generating \emph{answer rationales}, sequences of natural language and human-readable mathematical expressions that derive the final answer through a series of small steps. Although rationales do not explicitly specify programs, they provide a scaffolding for their structure via intermediate milestones. To evaluate our approach, we have created a new 100,000-sample dataset of questions, answers and rationales. Experimental results show that indirect supervision of program learning via answer rationales is a promising strategy for inducing arithmetic programs.
\end{abstract}

\section{Introduction}
Behaving intelligently often requires mathematical reasoning. Shopkeepers calculate change, tax, and sale prices; agriculturists calculate the proper amounts of fertilizers, pesticides, and water for their crops; and managers analyze productivity. Even determining whether you have enough money to pay for a list of items requires applying addition, multiplication, and comparison. Solving these tasks is challenging as it involves recognizing how goals, entities, and quantities in the real-world map onto a mathematical formalization, computing the solution, and mapping the solution back onto the world. As a proxy for the richness of the real world, a series of papers have used natural language specifications of algebraic word problems, and solved these by either learning to fill in templates that can be solved with equation solvers~\cite{DBLP:conf/emnlp/HosseiniHEK14,kushman-EtAl:2014:P14-1} or inferring and modeling operation sequences (programs) that lead to the final answer~\cite{Roy2015SolvingGA}.

\begin{figure}[t]
        {\fontsize{8.5}{9}\selectfont
        \hspace{-2mm}
        \begin{tabular}{|p{75mm}|}
        \hline
        \underline{\textbf{Problem 1}}:\\
        \textbf{Question}: Two trains running in opposite directions cross a man standing on the platform in 27 seconds and 17 seconds respectively and they cross each other in 23 seconds. The ratio of their speeds is:\\
		\textbf{Options}: A) 3/7$\ \ \ $B) 3/2$\ \ \ $C) 3/88$\ \ \ $D) 3/8$\ \ \ $E) 2/2 \\
		
        \textbf{Rationale}: Let the speeds of the two trains be x m/sec and y m/sec respectively. Then, length of the first train = 27x meters, and length of the second train = 17 y meters. (27x + 17y) / (x + y) = 23 $\rightarrow$ 27x + 17y = 23x + 23y $\rightarrow$ 4x = 6y $\rightarrow$ x/y = 3/2. \\
		\textbf{Correct Option}: B \\
        \hline
        
        \end{tabular}

		\hspace{-2mm}
        \begin{tabular}{|p{75mm}|}
        \hline
        \underline{\textbf{Problem 2}}:\\
        \textbf{Question}: From a pack of 52 cards, two cards are drawn together at random. What is the probability of both the cards being kings?\\
		\textbf{Options}: A) 2/1223$\ \ $ B) 1/122$\ \ $C) 1/221$\ \ $D) 3/1253$\ \ $E) 2/153 \\
		
        \textbf{Rationale}: Let s be the sample space.\\
		Then n(s) = 52C2 = 1326 \\
		E = event of getting 2 kings out of 4 \\
		n(E) = 4C2 = 6 \\
		P(E) = 6/1326 = 1/221 \\
		Answer is C \\
		\textbf{Correct Option}: C \\
        \hline
        \end{tabular}
		
		\hspace{-2mm}
        \begin{tabular}{|p{75mm}|}
        \hline
        \underline{\textbf{Problem 3}}:\\
        \textbf{Question}: For which of the following does $p(a)-p(b)=p(a-b)$ for all values of $a$ and $b$?\\
		\textbf{Options}:A) $p(x)=x^2$, B) $p(x)=x/2$, C) $p(x)=x+5$, D) $p(x)=2x−1$, E) $p(x)=|x|$ \\
		
        \textbf{Rationale}: To solve this easiest way is just put the value and see that if it equals or not. \\ 
		with option A. $p(a) = a^2$ and $p(b) = b^2$ \\
		so L.H.S = $a^2 - b^2$ \\
		and R.H.S = $(a-b)^2 \rightarrow a^2 + b^2 -2ab$. \\
		so L.H.S not equal to R.H.S \\
		with option B. $p(a) = a/2$ and $p(b) = b/2$ \\ 
		L.H.S = $a/2 - b/2 \rightarrow 1/2(a-b)$ \\
		R.H.S = $(a-b)/2$ \\ 
		so L.H.S = R.H.S which is the correct answer. \\ 
		answer:B\\
		
		\textbf{Correct Option}: B \\
        \hline
        
        \end{tabular}
        
        }
        
        \caption{Examples of solved math problems.}
\label{fig:examples}
\end{figure}

In this paper, we learn to solve algebraic word problems by inducing and modeling programs that generate not only the answer, but an \textbf{answer rationale}, a natural language explanation interspersed with algebraic expressions justifying the overall solution. Such rationales are what examiners require from students in order to demonstrate understanding of the problem solution; they play the very same role in our task. Not only do natural language rationales enhance model interpretability, but they provide a coarse guide to the structure of the arithmetic programs that must be executed. In fact the learner we propose (which relies on a heuristic search; \S\ref{sec:program_induction})
fails to solve this task without modeling the rationales---the search space is too unconstrained.

This work is thus related to models that can explain or rationalize their decisions~\citep{hendricks:2016,DBLP:journals/corr/HarrisonER17}. However, the use of rationales in this work is quite different from the role they play in most prior work, where interpretation models are trained to generate plausible sounding (but not necessarily accurate) post-hoc descriptions of the decision making process they used. In this work, the rationale is generated as a latent variable that gives rise to the answer---it is thus a more faithful representation of the steps used in computing the answer.

This paper makes three contributions. First, we have created a new dataset with more than 100,000 algebraic word problems that includes both answers and natural language answer rationales (\S\ref{sec:dataset}). Figure~\ref{fig:examples} illustrates three representative instances from the dataset. Second, we propose a sequence to sequence model that generates a sequence of instructions that, when executed, generates the rationale; only after this is the answer chosen (\S\ref{sec:model}). Since the target program is not given in the training data (most obviously, its specific form will depend on the operations that are supported by the program interpreter); the third contribution is thus a technique for inferring programs that generate a rationale and, ultimately, the answer. Even constrained by a text rationale, the search space of possible programs is quite large, and we employ a heuristic search to find plausible next steps to guide the search for programs (\S\ref{sec:program_induction}). Empirically, we are able to show that state-of-the-art sequence to sequence models are unable to perform above chance on this task, but that our model doubles the accuracy of the baseline~(\S\ref{sec:exp}).

\section{Dataset}
\label{sec:dataset}
We built a dataset\footnote{Available at \url{https://github.com/deepmind/AQuA}} with 100,000 problems with the annotations shown in Figure~\ref{fig:examples}. 
Each question is decomposed in four parts, two inputs and two outputs: the description of the problem, which we will denote as the \textbf{question}, and the possible (multiple choice) answer options, denoted as \textbf{options}. Our goal is to generate the description of the rationale used to reach the correct answer, denoted as \textbf{rationale} and the \textbf{correct option} label. Problem 1 illustrates an example of an algebra problem, which must be translated into an expression (i.e., $(27x + 17y) / (x + y) = 23$) and then the desired quantity $(x/y)$ solved for. Problem 2 is an example that could be solved by multi-step arithmetic operations proposed in~\cite{Roy2015SolvingGA}. Finally, Problem 3 describes a problem that is solved by testing each of the options, which has not been addressed in the past. 

\subsection{Construction}
We first collect a set of 34,202 seed problems that consist of multiple option math questions covering a broad range of topics and difficulty levels. Examples of exams with such problems include the GMAT (Graduate Management Admission Test) and GRE (General Test). Many websites contain example math questions in such exams, where the answer is supported by a rationale.

Next, we turned to crowdsourcing to generate new questions. We create a task where users are presented with a set of 5 questions from our seed dataset. Then, we ask the Turker to choose one of the questions and write a similar question. We also force the answers and rationale to differ from the original question in order to avoid paraphrases of the original question. Once again, we manually check a subset of the jobs for each Turker for quality control. The type of questions generated using this method vary. Some turkers propose small changes in the values of the questions (e.g., changing the equality $p(a)-p(b)=p(a-b)$ in Problem~3 to a different equality is a valid question, as long as the rationale and options are rewritten to reflect the change). We designate these as replica problems as the natural language used in the question and rationales tend to be only minimally unaltered. Others propose new problems in the same topic where the generated questions tend to differ more radically from existing ones. Some Turkers also copy math problems available on the web, and we define in the instructions that this is not allowed, as it will generate multiple copies of the same problem in the dataset if two or more Turkers copy from the same resource. These Turkers can be detected by checking the nearest neighbours within the collected datasets as problems obtained from online resources are frequently submitted by more than one Turker. Using this method, we obtained 70,318 additional questions.

\subsection{Statistics}

Descriptive statistics of the dataset is shown in Figure~\ref{stats}. In total, we collected 104,519 problems (34,202 seed problems and 70,318 crowdsourced problems). We removed 500 problems as heldout set (250 for development and 250 for testing). As replicas of the heldout problems may be present in the training set, these were removed manually by listing for each heldout instance the closest problems in the training set in terms of character-based Levenstein distance. After filtering, 100,949 problems remained in the training set.

We also show the average number of tokens (total number of tokens in the question, options and rationale) and the vocabulary size of the questions and rationales. Finally, we provide the same statistics exclusively for tokens that are numeric values and tokens that are not.

\begin{figure}[t]
  \begin{center}
    \centerline{\includegraphics[width=1.0\columnwidth,scale=0.22,clip=false,trim=0cm 0cm
    0cm 0cm]{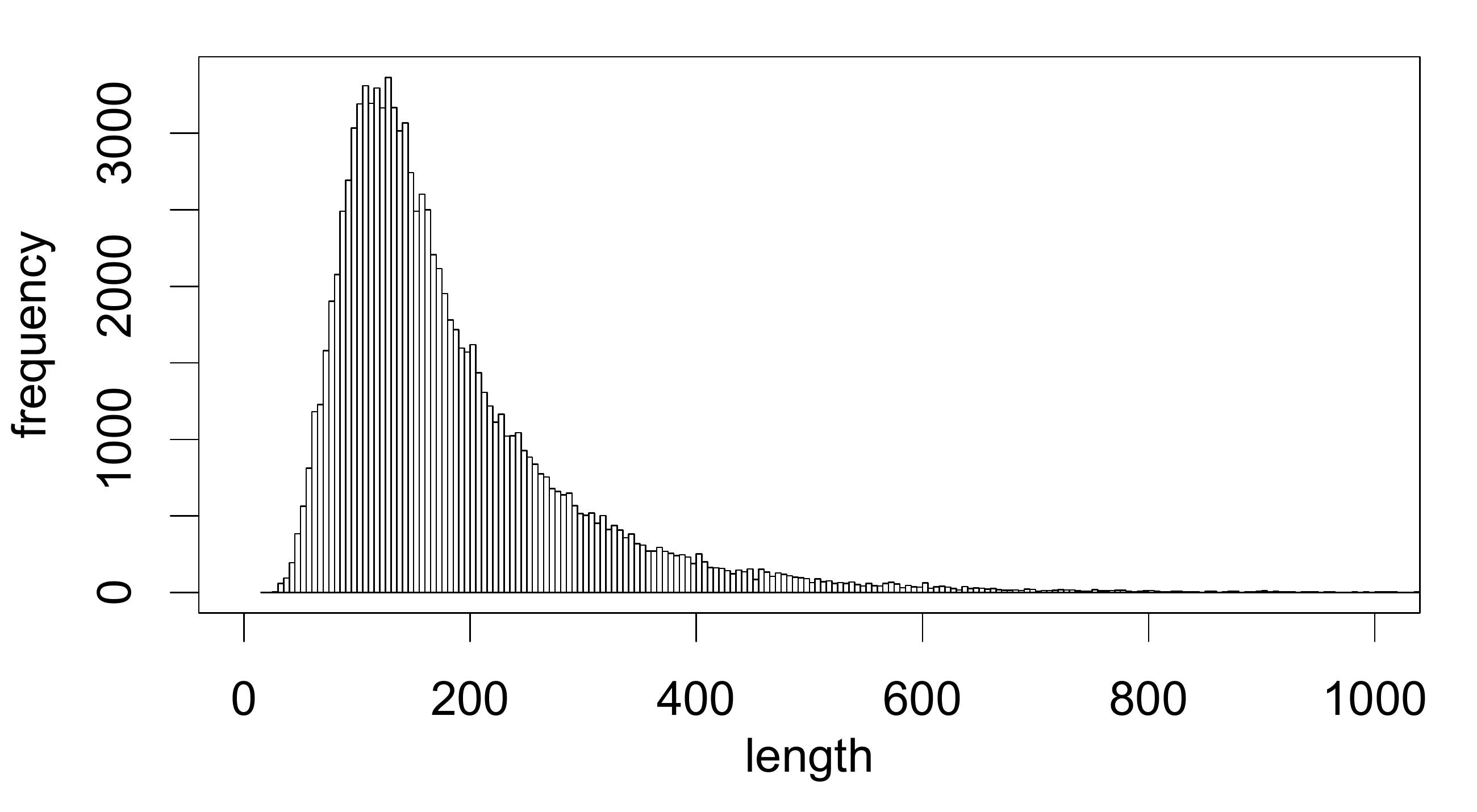}}
    \vspace{-0.5cm}
    \caption{Distribution of examples per length.}
    \label{histogram}
  \end{center}
\end{figure}

Figure~\ref{histogram} shows the distribution of examples based on the total number of tokens. We can see that most examples consist of 30 to 500 tokens, but there are also extremely long examples with more than 1000 tokens in our dataset.

\begin{table}[t]
\centering
\small
\begin{tabular}{|l|l|c|c|}
\hline
\multicolumn{2}{|c|}{} & \textbf{Question} & \textbf{Rationale} \\
\hline\hline
\multicolumn{2}{|c|}{Training Examples} & \multicolumn{2}{c|}{100,949}\\
\multicolumn{2}{|c|}{Dev Examples} & \multicolumn{2}{c|}{250}\\
\multicolumn{2}{|c|}{Test Examples} & \multicolumn{2}{c|}{250}\\
\hline\hline
\multirow{2}{*}{\textbf{Numeric}}&Average Length& 9.6 & 16.6 \\
&Vocab Size  & 21,009 & 14,745 \\
\hline
\multirow{2}{*}{\textbf{Non-Numeric}}&Average Length  & 67.8 & 89.1 \\
&Vocab Size & 17,849 & 25,034 \\
\hline\hline
\multirow{2}{*}{\textbf{All}}&Average Length  & 77.4 & 105.7 \\
&Vocab Size & 38,858 & 39,779 \\
\hline
\end{tabular}
\caption{Descriptive statistics of our dataset.}\label{stats}
\end{table}

\section{Model}
\label{sec:model}

Generating rationales for math problems is challenging as it requires models that learn to perform math operations at a finer granularity as each step within the solution must be explained. For instance, in Problem 1, the equation $(27x + 17y) / (x + y) = 23$ must be solved to obtain the answer. In previous work~\cite{kushman-EtAl:2014:P14-1}, this could be done by feeding the equation into an expression solver to obtain $x/y = 3/2$. However, this would skip the intermediate steps $27x + 17y = 23x + 23y$ and $4x = 6y$, which must also be generated in our problem. We propose a model that jointly learns to generate the text in the rationale, and to perform the math operations required to solve the problem. This is done by generating a program, containing both instructions that generate output and instructions that simply generate intermediate values used by following instructions.

\subsection{Problem Definition}

In traditional sequence to sequence models~\cite{DBLP:journals/corr/SutskeverVL14,DBLP:journals/corr/BahdanauCB14}, the goal is to predict the output sequence $\boldsymbol{y}=y_1,\ldots,y_{|\boldsymbol{y}|}$ from the input sequence $\boldsymbol{x}=x_1,\ldots,x_{|\boldsymbol{x}|}$, with lengths $|\boldsymbol{y}|$ and $|\boldsymbol{x}|$.

In our particular problem, we are given the problem and the set of options, and wish to predict the rationale and the correct option. We set $\boldsymbol{x}$ as the sequence of words in the problem, concatenated with words in each of the options separated by a special tag. Note that knowledge about the possible options is required as some problems are solved by the process of elimination or by testing each of the options (e.g. Problem 3). We wish to generate $\boldsymbol{y}$, which is the sequence of words in the rationale. We also append the correct option as the last word in $\boldsymbol{y}$, which is interpreted as the chosen option. For example, $\struct{y}$ in Problem 1 is ``Let the $\ldots$ = 3/2 . $\langle$EOR$\rangle$ B $\langle$EOS$\rangle$", whereas in Problem 2 it is ``Let s be $\dots$ Answer is C $\langle$EOR$\rangle$ C $\langle$EOS$\rangle$", where ``$\langle$EOS$\rangle$" is the end of sentence symbol and ``$\langle$EOR$\rangle$" is the end of rationale symbol.

\subsection{Generating Programs to Generate Rationales}

We wish to generate a latent sequence of \textbf{program instructions}, $\boldsymbol{z}=z_1,\ldots,z_{|\boldsymbol{z}|}$, with length ${|\boldsymbol{z}|}$, that will generate $\boldsymbol{y}$ when executed. 

We express $\boldsymbol{z}$ as a program that can access $\boldsymbol{x}$, $\boldsymbol{y}$, and the memory buffer $\boldsymbol{m}$. Upon finishing execution we expect that the sequence of output tokens to be placed in the output vector $\boldsymbol{y}$.

\begin{table}[t]
\centering
\small
\begin{tabular}{@{}r|l|l|l|l@{}}
\toprule
$i$ & $\boldsymbol{x}$ & $\boldsymbol{z}$ & $\boldsymbol{v}$ & $\boldsymbol{r}$\\
\midrule
1 & From & {\tt Id}(``Let") & \emph{Let} & $y_{1}$ \\
2 & a & {\tt Id}(``s") & \emph{s} & $y_{2}$ \\
3 & pack & {\tt Id}(``be") & \emph{be} & $y_{3}$ \\
4 & of & {\tt Id}(``the") & \emph{the} & $y_{4}$ \\
5 & 52 & {\tt Id}(``sample") & \emph{sample} & $y_{5}$ \\
6 & cards & {\tt Id}(``space") & \emph{space} & $y_{6}$ \\
7 & , & {\tt Id}(``.") & \emph{.} & $y_{7}$ \\
8 & two & {\tt Id}(``$\backslash$n") & \emph{$\backslash$n} & $y_{8}$ \\
9 & cards & {\tt Id}(``Then") & \emph{Then} & $y_{9}$ \\
10 & are & {\tt Id}(``n") & \emph{n} & $y_{10}$ \\
11 & drawn & {\tt Id}(``(") & \emph{(} & $y_{11}$ \\
12 & together & {\tt Id}(``s") & \emph{s} & $y_{12}$ \\
13 & at & {\tt Id}(``)") & \emph{)} & $y_{13}$ \\
14 & random & {\tt Id}(``=") & \emph{=} & $y_{14}$ \\
15 & . & {\tt Str\_to\_Float}($x_5$) & $\boldsymbol{52}$ & $\underline{m_{1}}$ \\
16 & What & {\tt Float\_to\_Str}($m_1$) & \emph{52} & $y_{15}$ \\
17 & is & {\tt Id}(``C") & \emph{C} & $y_{16}$ \\
18 & the & {\tt Id}(``2") & \emph{2} & $y_{17}$ \\
19 & probability & {\tt Id}(``=") & \emph{=} & $y_{18}$ \\
20 & of & {\tt Str\_to\_Float}($y_{17}$) & $\boldsymbol{2}$  & $\underline{m_{2}}$ \\
21 & both & {\tt Choose}($m_1$,$m_2$) & $\boldsymbol{1326}$ & $\underline{m_{3}}$ \\
22 & cards & {\tt Float\_to\_Str}($m_3$) & \emph{1326} & $y_{19}$ \\
23 & being & {\tt Id}(``E") & \emph{E} & $y_{20}$ \\
24 & kings & {\tt Id}(``=") & \emph{=} & $y_{21}$ \\
25 & ? & {\tt Id}(``event") & \emph{event} & $y_{22}$ \\
26 & $<$O$>$ & {\tt Id}(``of") & \emph{of} & $y_{23}$ \\
27 & A) & {\tt Id}(``getting") & \emph{getting} & $y_{24}$ \\
28 & 2/1223 & {\tt Id}(``2") & \emph{2} & $y_{25}$ \\
29 & $<$O$>$ & {\tt Id}(``kings") & \emph{kings} & $y_{26}$ \\
30 & B) & {\tt Id}(``out") & \emph{out} & $y_{27}$ \\
31 & 1/122 & {\tt Id}(``of") & \emph{of} & $y_{28}$\\
\ldots & \ldots & \ldots & \ldots & \ldots \\
$|\boldsymbol{z}|$& & {\tt Id}(``$\langle$EOS$\rangle$") & $\langle$\emph{EOS}$\rangle$ & $y_{|\boldsymbol{y}|}$ \\
\bottomrule
\end{tabular}
\caption{Example of a program $\boldsymbol{z}$ that would generate the output $\boldsymbol{y}$. In $\boldsymbol{v}$, \emph{italics} indicates string types; $\boldsymbol{bold}$ indicates float types. Refer to \S\ref{sec:instr} for description of variable names.\label{tab:code_example}}
\end{table}

Table~\ref{tab:code_example} illustrates an example of a sequence of instructions that would generate an excerpt from Problem 2, where columns $\boldsymbol{x}$, $\boldsymbol{z}$, $\boldsymbol{v}$, and $\boldsymbol{r}$ denote the input sequence, the instruction sequence (program), the values of executing the instruction, and where each value $v_i$ is written (i.e., either to the output or to the memory). In this example, instructions from indexes 1 to 14 simply fill each position with the observed output $y_1,\ldots,y_{14}$ with a string, where the \texttt{Id} operation simply returns its parameter without applying any operation. As such, running this operation is analogous to generating a word by sampling from a softmax over a vocabulary. However, instruction $z_{15}$ reads the input word $x_5$, 52, and applies the operation \texttt{Str\_to\_Float}, which converts the word 52 into a floating point number, and the same is done for instruction $z_{20}$, which reads a previously generated output word $y_{17}$. Unlike, instructions $z_{1},\ldots,z_{14}$, these operations write to the external memory $\boldsymbol{m}$, which stores intermediate values. 
A more sophisticated instruction---which shows some of the power of our model---is $z_{21}=\texttt{Choose}(m_1, m_2) \rightarrow m_3$ which evaluates ${m_1 \choose m_2}$ and
stores the result in $m_3$. 
This process repeats until the model generates the end-of-sentence symbol. The last token of the program as said previously must generate the correct option value, from ``A" to ``E".

By training a model to generate instructions that can manipulate existing tokens, the model benefits from the additional expressiveness needed to solve math problems within the generation process. In total we define 22 different operations, 13 of which are frequently used operations when solving math problems. These are: \texttt{Id}, \texttt{Add}, \texttt{Subtract}, \texttt{Multiply}, \texttt{Divide}, \texttt{Power}, \texttt{Log}, \texttt{Sqrt}, \texttt{Sine}, \texttt{Cosine}, \texttt{Tangent}, \texttt{Factorial}, and \texttt{Choose} (number of combinations). 
We also provide 2 operations to convert between \texttt{Radians} and \texttt{Degrees}, as these are needed for the sine, cosine and tangent operations. There are 6 operations that convert floating point numbers into strings and vice-versa. These include the \texttt{Str\_to\_Float} and \texttt{Float\_to\_Str} operations described previously, as well as operations which convert between floating point numbers and fractions, since in many math problems the answers are in the form ``3/4". For the same reason, an operation to convert between a floating point number and number grouped in thousands is also used (e.g. 1000000 to ``1,000,000'' or ``1.000.000").  Finally, we define an operation (\texttt{Check}) that given the input string, searches through the list of options and returns a string with the option index in \{``A'', ``B'', ``C'', ``D'', ``E''\}. If the input value does not match any of the options, or more than one option contains that value, it cannot be applied. For instance, in Problem 2, once the correct probability ``1/221" is generated, by applying the check operation to this number we can obtain correct option ``C".

\subsection{Generating and Executing Instructions}\label{sec:instr}

\ignore{
In a sequence to sequence model, we predict the probability of $\boldsymbol{y}$ given $\boldsymbol{x}$ as
$\log p(\boldsymbol{y} \mid \boldsymbol{x}) = \sum_{i} \log p(y_i \mid \boldsymbol{y}_{<i}, \boldsymbol{x})$, where
each token $y_i$ is predicted conditioned on the previously generated
sequence $\boldsymbol{y}_{<i}$ and input sequence $\boldsymbol{x}$. These probabilities are estimated with a
softmax over the vocabulary $\set{Y}$:
\begin{align*}
\label{char_softmax}
p(y_i \mid \boldsymbol{y}_{<i}, \boldsymbol{x}) = \underset{y_i \in \set{Y}}\softmax(\mathbf{h}_i)
\end{align*}
where $\mathbf{h}_i$ is the decoder RNN state at
timestamp $i$, which is modeled as $g(y_{i-1},\mathbf{h}_{i-1})$, and
$g(\cdot)$ is a recurrent update function, which generates the new state
$\mathbf{h}_i$ based on the previous token
$y_{i-1}$ and the previous state $\mathbf{h}_{t-1}$. In our work, we implement $g$ using a LSTM \cite{Hochreiter:1997:LSM:1246443.1246450}. 
The initial state of the decoder $\mathbf{h}_0$ is the task state of the encoder, which is implemented as an LSTM chain over the input sequence $\boldsymbol{x}$ with hidden states $\mathbf{u}_0,\ldots,\mathbf{u}_{|x|}$.
}

In our model, programs consist of sequences of instructions, $\boldsymbol{z}$. We turn now to how we model each $z_i$, conditional on the text program specification, and the program's history. The instruction $z_i$ is a tuple consisting of an operation ($o_i$), an ordered sequence of its arguments ($\boldsymbol{a}_i$), and a decision about where its results will be placed ($r_i)$ (is it appended in the output $\boldsymbol{y}$ or in a memory buffer $\boldsymbol{m}$?), and the result of applying the operation to its arguments ($v_i$). That is, $z_i=(o_i,r_i,\boldsymbol{a}_{i},v_i)$. 

Formally, $o_i$ is an element of the pre-specified set of operations $\mathcal{O}$, which contains, for example $\texttt{add}$, $\texttt{div}$, $\texttt{Str\_to\_Float}$, etc. The number of arguments required by $o_i$ is given by $\textit{argc}(o_i)$, e.g., $\textit{argc}(\texttt{add})=2$ and $\textit{argc}(\texttt{log})=1$. The arguments are $\boldsymbol{a}_i = a_{i,1},\ldots,a_{i,\textit{argc}(o_i)}$. An instruction will generate a return value $v_i$ upon execution, which will either be placed in the output $\boldsymbol{y}$ or hidden. This decision is controlled by $r_i$. We define the instruction probability as:
\begin{align*}
p(o_i,\boldsymbol{a}_i, r_i, &v_i \mid \boldsymbol{z}_{<i}, \boldsymbol{x}, \boldsymbol{y}, \boldsymbol{m}) = \\
&p(o_i \mid \boldsymbol{z}_{<i},\boldsymbol{x}) \times p(r_i \mid \boldsymbol{z}_{<i},\boldsymbol{x},o_i) \times \\
& \qquad \prod_{j=1}^{\textit{argc}(o_i)} p(a_{i,j} \mid \boldsymbol{z}_{<i},\boldsymbol{x},o_i,\boldsymbol{m},\boldsymbol{y}) \times \\
&\qquad [v_i = \text{apply}(o_i,\boldsymbol{a})],
\end{align*}
where $[p]$ evaluates to 1 if $p$ is true and 0 otherwise, and $\text{apply}(f,\boldsymbol{x})$ evaluates the operation $f$ with arguments $\boldsymbol{x}$. Note that the apply function is not learned, but pre-defined.

\begin{figure}[t]
\centering
\includegraphics[scale=0.277,clip=false,trim=1cm 3cm
    1cm 5cm]{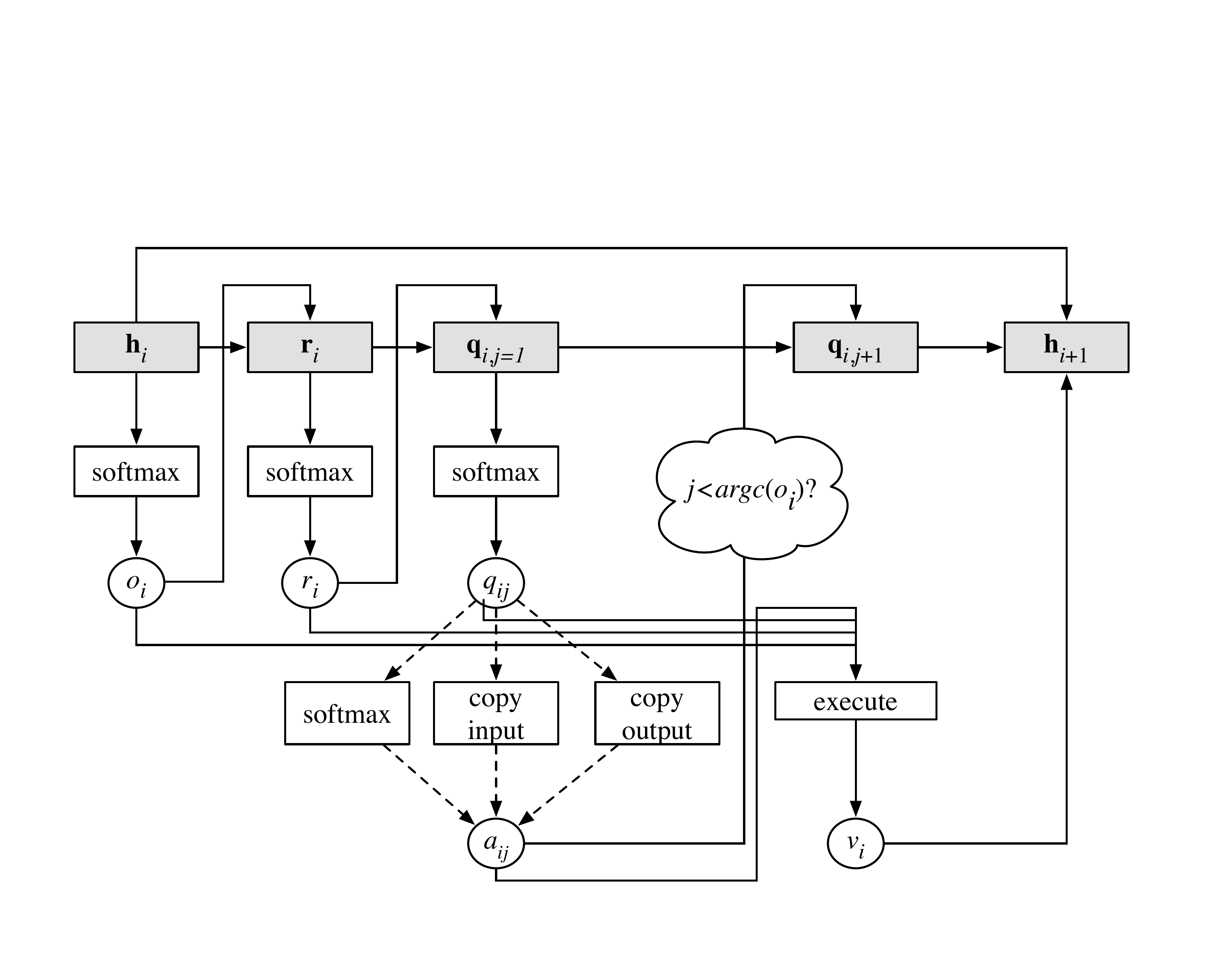}
    \vspace{-.4cm}\caption{Illustration of the generation process of a single instruction tuple at timestamp $i$.}
    \label{input_img}
\end{figure}

The network used to generate an instruction at a given timestamp $i$ is illustrated in Figure~\ref{input_img}. 
We first use the recurrent state $\mathbf{h}_i$ to generate $p(o_i\mid \boldsymbol{z}_{<i},\boldsymbol{x})=\underset{o_i \in \set{O}}\softmax(\mathbf{h}_i)$, using a softmax over the set of available operations $\set{O}$.

In order to predict $r_i$, we generate a new hidden state $\mathbf{r}_i$, which is a function of the current program context $\mathbf{h}_i$, and an embedding of the current predicted operation, $o_i$. 
As the output can either be placed in the memory  $\boldsymbol{m}$ or the output $\boldsymbol{y}$, we 
compute the probability $p(r_i = \textsc{output} \mid \boldsymbol{z}_{<i},\boldsymbol{x},o_i)=\sigma (\mathbf{r}_{i} \cdot \vect{w}_r +b_r)$, where $\sigma$ is the logistic 
sigmoid function. 
If $r_i=\textsc{output}$, $v_i$ is appended to the output $\boldsymbol{y}$; otherwise it is appended to the memory $\boldsymbol{m}$.

Once we generate $r_i$, we must predict 
$\boldsymbol{a}_i$, the $\textit{argc}(o_i)$-length sequence of arguments that operation $o_i$ requires.
The $j$th argument $a_{i,j}$ can be either generated from a softmax over the vocabulary, copied from the input vector $\boldsymbol{x}$, or copied from previously generated values in the output $\boldsymbol{y}$ or memory $\boldsymbol{m}$. This decision is modeled using a 
latent predictor network~\cite{DBLP:journals/corr/LingGHKSWB16}, where the control over which method used to generate $a_{i,j}$ is governed by a latent variable $q_{i,j} \in \{\textsc{softmax}, \textsc{copy-input}, \textsc{copy-output}\}$.
Similar to when predicting $r_i$, 
in order to make this choice, 
we also generate a new hidden state for each argument slot $j$,
denoted by $\vect{q}_{i,j}$ with an LSTM, feeding the previous argument in at each time step, and initializing it with $\mathbf{r}_i$ and by reading the predicted value of the output $r_i$.
\begin{itemizesquish}
\item If $q_{i,j} = \textsc{softmax}$, $a_{i,j}$ is generated by sampling from a softmax over the vocabulary $\set{Y}$,
\begin{align*}
\small
p(a_{i,j}&\mid q_{i,j}=\textsc{softmax})=\underset{a_{i,j} \in \set{Y}}\softmax(\mathbf{q}_{i,j}).
\end{align*}
This corresponds to a case where a string is used as argument (e.g. $y_1$=``Let"). 
\item If $q_{i,j} = \textsc{copy-input}$, $a_{i,j}$ is obtained by copying an element from the input vector with a pointer network~\cite{NIPS2015_5866} over input words $x_1,\ldots,x_{|\boldsymbol{x}|}$, represented by their encoder LSTM state $\mathbf{u}_1,\ldots,\mathbf{u}_{|\boldsymbol{x}|}$. As such, we compute the probability distribution over input words as:
\begin{align}
\label{copy-mec}
p(a_{i,j} \mid q_{i,j} =& \textsc{copy-input}) = \\ 
&\underset{a_{i,j} \in x_1,\ldots,x_{|\boldsymbol{x}|}}\softmax\left(f(\mathbf{u}_{a_{i,j}},\mathbf{q}_{i,j})\right)\nonumber
\end{align}
Function $f$ computes the affinity of each token $x_{a_{i,j}}$ and the current
output context $\mathbf{q}_{i,j}$. A common implementation of $f$, which we follow, is to apply a
linear projection from [$\mathbf{u}_{a_{i,j}};\mathbf{q}_{i,j}]$ into a fixed size vector (where $[\mathbf{u};\mathbf{v}]$ is vector concatenation), followed by a $\tanh$ and a linear projection into a single value.
\item If $q_{i,j} = \textsc{copy-output}$, the model copies from either the output $\boldsymbol{y}$ or the memory $\boldsymbol{m}$. This is equivalent to finding the instruction $z_i$, where the value was generated. Once again, we define a pointer network that points to the output instructions and define the distribution over previously generated instructions as:
\begin{align*}
p(a_{i,j} \mid q_{i,j} =& \textsc{copy-output}) =\\ 
&\underset{a_{i,j} \in z_1,\ldots,z_{i-1}}\softmax\left(f(\mathbf{h}_{a_{i,j}},\mathbf{q}_{i,j})\right)
\end{align*}
Here, the affinity is computed using the decoder state $\mathbf{h}_{a_{i,j}}$ and the current state $\mathbf{q}_{i,j}$. 
\end{itemizesquish}

Finally, we embed the argument $a_{i,j}$\footnote{
The embeddings of a given argument $a_{i,j}$ and the return value $v_{i}$ are obtained with a lookup table embedding and two flags indicating whether it is a string and whether it is a float. Furthermore, if the the value is a float we also add its numeric value as a feature. 
}
and the state $\mathbf{q}_{i,j}$ to generate the next state $\mathbf{q}_{i,j+1}$. Once all arguments for $o_i$ are generated, the operation is executed to obtain $v_i$. 
Then, the embedding of $v_i$, the final state of the instruction $\mathbf{q}_{i,|\boldsymbol{a}_i|}$ and the previous state $\mathbf{h}_i$ are used to generate the state at the next timestamp $\mathbf{h}_{i+1}$.

\section{Inducing Programs while Learning}
\label{sec:program_induction}

The set of instructions $\boldsymbol{z}$ that will generate $\boldsymbol{y}$ is unobserved. Thus, given $\boldsymbol{x}$ we optimize the marginal probability function:

\begin{align*}
p(\boldsymbol{y}\mid \boldsymbol{x}) = \sum_{\boldsymbol{z} \in \set{Z}} p(\boldsymbol{y}\mid \boldsymbol{z}) p(\boldsymbol{z} \mid \boldsymbol{x}) = \sum_{\boldsymbol{z} \in \set{Z}(y)} p(\boldsymbol{z} \mid \boldsymbol{x}),
\end{align*}
where $p(\boldsymbol{y} \mid \boldsymbol{z})$ is the Kronecker delta function $\delta_{e(\boldsymbol{z}),\boldsymbol{y}}$, which is 1 if the execution of $\boldsymbol{z}$, denoted as $e(\boldsymbol{z})$, generates $y$ and 0 otherwise. Thus, we can redefine $p(\boldsymbol{y}|\boldsymbol{x})$, the marginal over all programs $\set{Z}$, as a marginal over programs that would generate $\boldsymbol{y}$, defined as $\set{Z}(\boldsymbol{y})$. As marginalizing over $\boldsymbol{z}\in \set{Z}(\boldsymbol{y})$ is intractable, we approximate the marginal by generating samples from our model. Denote the set of samples that
are generated by $\hat{\set{Z}}(\boldsymbol{y})$. We maximize $\sum{\boldsymbol{z} \in \hat{\set{Z}}(\boldsymbol{y})} p(\boldsymbol{z}|\boldsymbol{x})$.

However, generating programs that generate $\boldsymbol{y}$ is not trivial, as randomly sampling from the RNN distribution over instructions at each timestamp is unlikely to generate a sequence $\boldsymbol{z}\in \set{Z}(\boldsymbol{y})$.

This is analogous to the question answering work in~\newcite{DBLP:journals/corr/LiangBLFL16}, where the query that generates the correct answer must be found during inference, and training proved to be difficult without supervision. In~\newcite{Roy2015SolvingGA} this problem is also addressed by adding prior knowledge to constrain the exponential space. 

In our work, we leverage the fact that we are generating rationales, where there is a sense of progression within the rationale. That is, we assume that the rationale solves the problem step by step. For instance, in Problem 2, the rationale first describes the number of combinations of two cards in a deck of 52 cards, then describes the number of combinations of  two kings, and finally computes the probability of drawing two kings. Thus, while generating the final answer without the rationale requires a long sequence of latent instructions, generating each of the tokens of the rationale requires far less operations. 

More formally, given the 
sequence $z_1,\dots,z_{i-1}$ generated so far, and the possible values for $z_i$ 
given by the network, denoted $\set{Z}_i$, 
we wish to filter $\set{Z}_i$ to $\set{Z}_i(y_k)$, which denotes a set of possible options that contain at least one path capable of generating the next token at index $k$. Finding the set $\set{Z}_i(y_k)$ is achieved by testing all combinations of instructions that are possible with at most one level of indirection, and keeping those that can generate $y_k$. This means that the model can only generate one intermediate value in memory (not including the operations that convert strings into floating point values and vice-versa). 

\paragraph{Decoding.}
During decoding we find the most likely sequence of instructions $\struct{z}$ given $\struct{x}$, which can be performed with a stack-based decoder. However, it is important to refer that each generated instruction $z_i=(o_i,r_i,a_{i,1},\ldots,a_{i,|a_{i}|},v_i)$ must be executed to obtain $v_i$. To avoid generating unexecutable code---e.g., log(0)---each hypothesis instruction is executed and removed if an error occurs. Finally, once the ``$\langle$EOR$\rangle$" tag is generated, we only allow instructions that would generate one of the option ``A" to ``E" to be generated, which guarantees that one of the options is chosen.

\section{Staged Back-propagation}
As it is shown in Figure~\ref{histogram}, math rationales with more than 200 tokens are not uncommon, and with additional intermediate instructions, the size $\boldsymbol{z}$ can easily exceed 400. This poses a practical challenge for training the model.

For both the attention and copy mechanisms, for each instruction $z_i$, the model needs to compute the probability distribution between all the attendable units $\boldsymbol{c}$ conditioned on the previous state $\mathbf{h}_{i-1}$. For the attention model and input copy mechanisms, $\boldsymbol{c}=\boldsymbol{x_{0,i-1}}$ and for the output copy mechanism $\boldsymbol{c}=\boldsymbol{z}$. These operations generally involve an exponential number of matrix multiplications as the size of $\boldsymbol{c}$ and $\boldsymbol{z}$ grows. For instance, during the computation of the probabilities for the input copy mechanism in Equation~\ref{copy-mec}, the affinity function $f$ between the current context $\boldsymbol{q}$ and a given input $\boldsymbol{u}_k$ is generally implemented by projecting $\boldsymbol{u}$ and $\boldsymbol{q}$ into a single vector followed by a non-linearity, which is projected into a single affinity value. Thus, for each possible input $\boldsymbol{u}$, 3 matrix multiplications must be performed. Furthermore, for RNN unrolling, parameters and intermediate outputs for these operations must be replicated for each timestamp. Thus, as $\boldsymbol{z}$ becomes larger the attention and copy mechanisms quickly become a memory bottleneck as the computation graph becomes too large to fit on the GPU. In contrast, the sequence-to-sequence model proposed in ~\cite{DBLP:journals/corr/SutskeverVL14}, does not suffer from these issues as each timestamp is dependent only on the previous state $\mathbf{h}_{i-1}$.

To deal with this, we use a training method we call \textbf{staged back-propagation} which saves memory by considering slices of $K$ tokens in $\boldsymbol{z}$, rather than the full sequence. That is, to train on a mini-batch where $|z|=300$ with $K=100$, we would actually train on 3 mini-batches, where the first batch would optimize for the first $\boldsymbol{z}_{1:100}$, the second for $\boldsymbol{z}_{101:200}$ and the third for $\boldsymbol{z}_{201:300}$. The advantage of this method is that memory intensive operations, such as attention and the copy mechanism, only need to be unrolled for $K$ steps, and $K$ can be adjusted so that the computation graph fits in memory. 

However, unlike truncated back-propagation for language modeling, where context outside the scope of $K$ is ignored, sequence-to-sequence models require global context. Thus, the sequence of states $\boldsymbol{h}$ is still built for the whole sequence $\boldsymbol{z}$. Afterwards, we obtain a slice $\boldsymbol{h}_{j:j+K}$, and compute the attention vector.\footnote{This modeling strategy is sometimes known as late fusion, as the attention vector is not used for state propagation, it is incorporated ``later''.} Finally, the prediction of the instruction is conditioned on the LSTM state and the attention vector.

\section{Experiments}
\label{sec:exp}
We apply our model to the task of generating rationales for solutions to math problems, evaluating it on both the quality of the rationale and the ability of the model to obtain correct answers. 


\subsection{Baselines}
As the baseline we use the attention-based sequence to sequence model proposed by~\newcite{DBLP:journals/corr/BahdanauCB14}, and proposed augmentations, allowing it to copy from the input~\cite{DBLP:journals/corr/LingGHKSWB16} and from the output~\cite{DBLP:journals/corr/MerityXBS16}.

\subsection{Hyperparameters}
We used a two-layer LSTM with a hidden size of $H=200$, and word embeddings with size 200. The number of levels that the graph $\mathcal{G}$ is expanded during sampling $D$ is set to 5. Decoding is performed with a beam of 200. As for the vocabulary of the softmax and embeddings, we keep the most frequent 20,000 word types, and replace the rest of the words with an unknown token. During training, the model only learns to predict a word as an unknown token, when there is no other alternative to generate the word.

\subsection{Evaluation Metrics}

The evaluation of the rationales is performed with average sentence level perplexity and BLEU-4~\cite{Papineni:2002:BMA:1073083.1073135}. When a model cannot generate a token for perplexity computation, we predict unknown token. This benefits the baselines as they are less expressive. As the perplexity of our model is dependent on the latent program that is generated, we force decode our model to generate the rationale, while maximizing the probability of the program. This is analogous to the method used to obtain sample programs described in Section~\ref{sec:program_induction}, but we choose the most likely instructions at each timestamp instead of sampling. Finally, the correctness of the answer is evaluated by computing the percentage of the questions, where the chosen option matches the correct one.

\subsection{Results}

The test set results, evaluated on perplexity, BLEU, and accuracy, are presented in Table~\ref{tab:results}.

\begin{table}[t]
\centering
\begin{tabular}{|l|r|r|r|}
\hline
Model & Perplexity & BLEU & Accuracy\\
\hline
Seq2Seq & 524.7 & 8.57 & 20.8\\
+Copy Input & 46.8 & 21.3 & 20.4\\
+Copy Output & 45.9 & 20.6 & 20.2\\
\hline
Our Model & \textbf{28.5} & \textbf{27.2} & \textbf{36.4}\\
\hline
\end{tabular}
\caption{Results over the test set measured in Perplexity, BLEU and Accuracy.}\label{tab:results}
\end{table}

\begin{figure*}[t]
\vspace{-0.8cm}
\centering
\includegraphics[scale=0.6]{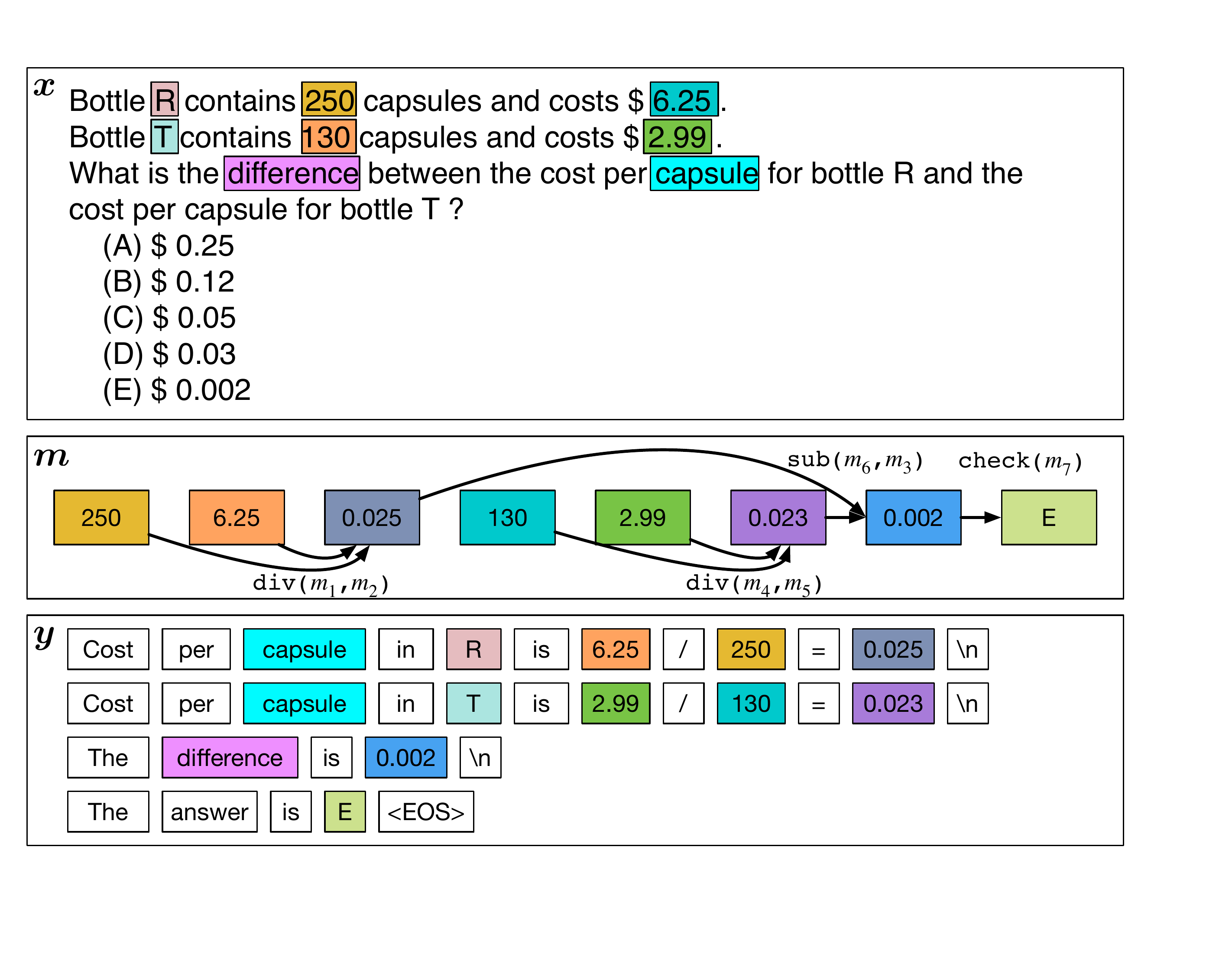}
    \vspace{-2cm}\caption{Illustration of the most likely latent program inferred by our algorithm to explain a held-out question-rationale pair.}
    \label{example}
\end{figure*}

\paragraph{Perplexity.} In terms of perplexity, we observe that the regular sequence to sequence model fares poorly on this dataset, as the model requires the generation of many values that tend to be sparse. Adding an input copy mechanism greatly improves the perplexity as it allows the generation process to use values that were mentioned in the question. The output copying mechanism improves perplexity slightly over the input copy mechanism, as many values are repeated after their first occurrence. For instance, in Problem 2, the value ``1326" is used twice, so even though the model cannot generate it easily in the first occurrence, the second one can simply be generated by copying the first one. We can observe that our model yields significant improvements over the baselines, demonstrating that the ability to generate new values by algebraic manipulation is essential in this task. An example of a program that is inferred is shown in Figure ~\ref{example}. The graph was generated by finding the most likely program $\boldsymbol{z}$ that generates $\boldsymbol{y}$. Each node isolates a value in $\boldsymbol{x}$, $\boldsymbol{m}$, or $\boldsymbol{y}$, where arrows indicate an operation executed with the outgoing nodes as arguments and incoming node as the return of the operation. For simplicity, operations that copy or convert values (e.g. from string to float) were not included, but nodes that were copied/converted share the same color. Examples of tokens where our model can obtain the perplexity reduction are the values ``0.025", ``0.023", ``0.002" and finally the answer ``E" , as these cannot be copied from the input or output.

\paragraph{BLEU.}
We observe that the regular sequence to sequence model achieves a low BLEU score. In fact, due to the high perplexities the model generates very short rationales, which frequently consist of segments similar to ``Answer should be D", as most rationales end with similar statements. By applying the copy mechanism the BLEU score improves substantially, as the model can define the variables that are used in the rationale. Interestingly, the output copy mechanism adds no further improvement in the perplexity evaluation. This is because during decoding all values that can be copied from the output are values that could have been generated by the model either from the softmax or the input copy mechanism. As such, adding an output copying mechanism adds little to the expressiveness of the model during decoding.

Finally, our model can achieve the highest BLEU score as it has the mechanism to generate the intermediate and final values in the rationale.

\paragraph{Accuracy.}
In terms of accuracy, we see that all baseline models obtain values close to chance (20\%), indicating that they are completely unable to solve the problem. In contrast, we see that our model can solve problems at a rate that is significantly higher than chance, demonstrating the value of our program-driven approach, and its ability to learn to generate programs.


In general, the problems we solve correctly correspond to simple problems that can be solved in one or two operations. Examples include questions such as ``Billy cut up each cake into 10 slices, and ended up with 120 slices altogether. How many cakes did she cut up? A) 9 B) 7 C) 12 D) 14 E) 16", which can be solved in a single step. In this case, our model predicts ``120 / 10 = 12 cakes. Answer is C" as the rationale, which is reasonable.

\subsection{Discussion.}
While we show that our model can outperform the models built up to date, generating complex rationales as those shown in Figure~\ref{fig:examples} correctly is still an unsolved problem, as each additional step adds complexity to the problem both during inference and decoding. Yet, this is the first result showing that it is possible to solve math problems in such a manner, and we believe this modeling approach and dataset will drive work on this problem.

\section{Related Work}
Extensive efforts have been made in the domain of math problem solving~\cite{DBLP:conf/emnlp/HosseiniHEK14,kushman-EtAl:2014:P14-1,Roy2015SolvingGA}, which aim at obtaining the correct answer to a given math problem. Other work has focused on learning to map math expressions into formal languages~\cite{DBLP:journals/corr/RoyUR16}. We aim to generate natural language rationales, where the bindings between variables and the problem solving approach are mixed into a single generative model that attempts to solve the problem while explaining the approach taken.

Our approach is strongly tied with the work on sequence to sequence transduction using the encoder-decoder paradigm~\cite{DBLP:journals/corr/SutskeverVL14,DBLP:journals/corr/BahdanauCB14,kalchbrenner2013recurrent}, and inherits ideas from the extensive literature on semantic parsing~\cite{Jones:2012:SPB:2390524.2390593,BerantCFL13,andreas-vlachos-clark:2013:Short,quirk:acl15,DBLP:journals/corr/LiangBLFL16,neelakantan:2016} and program generation~\cite{DBLP:journals/corr/ReedF15,Graves_Nature2016}, namely, the usage of an external memory, the application of different operators over values in the memory and the copying of stored values into the output sequence. 

Providing textual explanations for classification decisions has begun to receive attention, as part of increased interest in creating models whose decisions can be interpreted. \newcite{lei:2016}, jointly modeled both a classification decision, and the selection of the most relevant subsection of a document for making the classification decision. 
\newcite{hendricks:2016} generate textual explanations for visual classification problems, but in contrast to our model, they first generate an answer, and then, conditional on the answer, generate an explanation. This effectively creates a post-hoc justification for a classification decision rather than a program for deducing an answer. These papers, like ours, have jointly modeled rationales and answer predictions; however, we are the first to use rationales to guide program induction.

\section{Conclusion}
In this work, we addressed the problem of generating rationales for math problems, where the task is to not only obtain the correct answer of the problem, but also generate a description of the method used to solve the problem. To this end, we collect 100,000 question and rationale pairs, and propose a model that can generate natural language and perform arithmetic operations in the same decoding process. Experiments show that our method outperforms existing neural models, in both the fluency of the rationales that are generated and the ability to solve the problem.

\bibliography{paper}
\bibliographystyle{acl_natbib}

\appendix

\end{document}